\documentclass{article}

\usepackage[table]{xcolor}
\PassOptionsToPackage{numbers}{natbib}
\usepackage[preprint]{neurips_2026}

\usepackage[utf8]{inputenc} % allow utf-8 input
\usepackage{float}
\usepackage[T1]{fontenc}    % use 8-bit T1 fonts
\usepackage{hyperref}       % hyperlinks
\usepackage{url}            % simple URL typesetting
\usepackage{amsfonts}       % blackboard math symbols
\usepackage{nicefrac}       % compact symbols for 1/2, etc.
\usepackage{microtype}      % microtypography
\usepackage{graphicx}
\usepackage{amssymb}
\usepackage{tablefootnote}
\usepackage{booktabs,caption,fixltx2e}
\usepackage[flushleft]{threeparttable}
\usepackage[accsupp]{axessibility}
\usepackage{multicol}
\usepackage{multirow}
\usepackage{listings}
\usepackage{pifont}
\usepackage{wrapfig}
\usepackage{xcolor}         % colors
\usepackage{footnote}
\makesavenoteenv{tabular}

\usepackage{graphicx}
\usepackage{booktabs}
\usepackage{amssymb}
\usepackage{microtype}
\usepackage{tcolorbox}
\tcbuselibrary{breakable, skins}
\usepackage{enumitem}
\usepackage[accsupp]{axessibility}  % Improves PDF readability for those with disabilities.
\usepackage{listings}
\definecolor{jsonkey}{RGB}{0, 102, 204}      % 蓝色：key
\definecolor{jsonstring}{RGB}{163, 21, 21}   % 深红：string value
\definecolor{jsonnumber}{RGB}{9, 134, 88}    % 绿色：number

\lstdefinelanguage{json}{
  basicstyle=\ttfamily\small,
  morestring=[b]",
  stringstyle=\color{jsonstring},
  literate=
    *{0}{{{\color{jsonnumber}0}}}{1}
     {1}{{{\color{jsonnumber}1}}}{1}
     {2}{{{\color{jsonnumber}2}}}{1}
     {3}{{{\color{jsonnumber}3}}}{1}
     {4}{{{\color{jsonnumber}4}}}{1}
     {5}{{{\color{jsonnumber}5}}}{1}
     {6}{{{\color{jsonnumber}6}}}{1}
     {7}{{{\color{jsonnumber}7}}}{1}
     {8}{{{\color{jsonnumber}8}}}{1}
     {9}{{{\color{jsonnumber}9}}}{1}
}

\usepackage[T1]{fontenc}
\usepackage[utf8]{inputenc}

\title{Seeing Without Exposing: Adaptive Privacy Control for Open-World, Context-Hungry MLLMs}
\author{%
  Siyuan Xu\thanks{Equal contribution.} \\
  City University of \\
  Hong Kong \\
  \texttt{siyuanxu333@gmail.com} \\
  \And
  Yibing Liu\footnotemark[1] \\
  City University of \\
  Hong Kong \\
  \texttt{lyibing112@gmail.com} \\
  \And
  Peilin Chen \\
  City University of \\
  Hong Kong \\
  \texttt{plchen3@cityu.edu.hk} \\
  \AND
  Yung-Hui Li \\
  Hon Hai \\
   Research Institute \\
  \texttt{yunghui.li@foxconn.com} \\
  \And
  Shiqi Wang \\
  City University of \\
  Hong Kong \\
  \texttt{shiqwang@cityu.edu.hk} \\
  \And
  Sam Kwong \\
  Lingnan  \\ 
  University \\
  \texttt{samkwong@ln.edu.hk} \\
}

\begin{document}

\maketitle

% \begin{center}
% \includegraphics[width=\textwidth]{figs/teaser.pdf}

% % \vspace{1mm}

% \end{center}
% \small

% \begin{figure}
%     \centering
%     % \vspace{1.5mm}
%     \includegraphics[width=1.\textwidth]{figs/teaser.pdf}
%     % \vspace{0.5mm}
%     \caption{\textbf{Privacy-Informed Editing (PIE)} defines a new privacy protection paradigm for the MLLM era. Unlike existing privacy protection approaches, PIE preserves content coherence and high perceptual quality while explicitly maintaining information beneficial for \textit{downstream tasks} and \textit{personalized} user requirements across diverse private inputs.}
%     \label{fig:teaser}
% \end{figure}

\begin{abstract}
    Multimodal large language models (MLLMs) have raised new privacy challenges. On the data side, user-provided inputs often include unpredictable sensitive information; while on the downstream task side, model reasoning depends on rich visual context that may itself be privacy-sensitive.
    Existing privacy protection methods, however, rely on predefined sensitive categories and fixed obfuscation strategies, struggling to tackle such challenges in MLLMs.
    To address this dilemma, we propose Anchored Privacy Drifting (APD), a training-free method that drifts privacy-sensitive elements toward semantically equivalent alternatives while anchoring contextual cues to the source image.
    To systematically evaluate this dual objective of privacy protection and contextual preservation, we introduce AdaptShield, a comprehensive benchmark covering 22 privacy categories, which combines conventional privacy metrics with MLLM-based assessments of contextual utility. 
    Extensive experiments show that our method achieves balanced improvements in both privacy sanitization and content retention, with average gains of 10.4\% on textual categories and 8.5\% under MLLM-based evaluation across four MLLM series, \textit{i.e.}, Qwen2.5, Qwen3, InternVL3, and InternVL3.5.
\end{abstract}

\section{Introduction}
\label{sec:intro}

Visual privacy protection for Multimodal Large Language Models (MLLMs) has seen intensified concerns recently~\cite{zhang2024multip2a,zhang2024multitrust,safe_llava_2025,gu2024mllmguardmultidimensionalsafetyevaluation}. Unlike conventional definitions of privacy that emphasize limited entities (\textit{e.g.,} faces), visual privacy in MLLMs presents a distinct \textit{user-tailored} paradigm -- what constitutes sensitive information varies across users, contexts, and intents. 
A scene, object, or gesture that appears innocuous to one user may carry private or revealing meanings to another~\cite{phantom2024, zhang2024multip2a,zhang2025geo,luo2025doxinglensrevealinglocationrelated,mendes-etal-2024-granular,vip,cao2024facedeidentificationstateoftheartmethods,MEDEN2023104678}. 
This personalized nature challenges the one-size-fits-all assumption 
of traditional anonymization, motivating privacy formulations that 
incorporate user-provided specifications such as region masks or 
explicit concealment intents. Beyond concealment, since protected images are further consumed by models for multimodal understanding and reasoning, high-fidelity protection is equally critical in MLLM settings.  Severe distortions to scene layout, object relationships, or contextual cues may make an image privacy-safe but ineffective for downstream tasks.
% \vspace{2pt}
% \\

 However, existing approaches remain largely polarized. Specialized methods such as facial de-identification~\cite{nullface,face_simple,falco,riddle} achieve fine-grained concealment but rely on domain-specific priors~\cite{psp,e4e}, limiting scalability to broader privacy scenarios. Universal approaches instead adopt coarse obfuscation strategies~\cite{li2017effectiveness,fan2018image} that suppress sensitive content at the cost of destroying contextual and structural information, rendering protected images unreliable for downstream MLLM understanding and reasoning. 
Consequently, there still lacks a unified framework that can adaptively balance privacy concealment and content preservation under diverse and personalized privacy requirements.
% \vspace{2pt}
% \\

To this end, we propose \textit{Anchored Privacy Drifting (APD)}, a training-free framework that generates a privacy-safe version of the source image with transformed private content, while preserving the overall visual fidelity. APD operates in a shared multimodal latent space and steers the denoising trajectory with two directional signals to control where the generated image ultimately lands. Under high-level semantic guidance, the latent representation drifts toward a privacy-safe region whose private specifics naturally diverge from the source through inherent semantic stochasticity. In parallel, a source-derived direction, computed directly between the source and the current state, anchors the trajectory against excessive deviation from the original context. By adaptively balancing these two directions across spatial regions and generation steps, APD steers the trajectory toward an endpoint that is structurally faithful to the source yet semantically redirected away from private specifics, achieving controllable privacy concealment with minimal contextual degradation.
\begin{figure}[t]
    \vspace{-25pt}
    \centering
    \includegraphics[width=\linewidth]{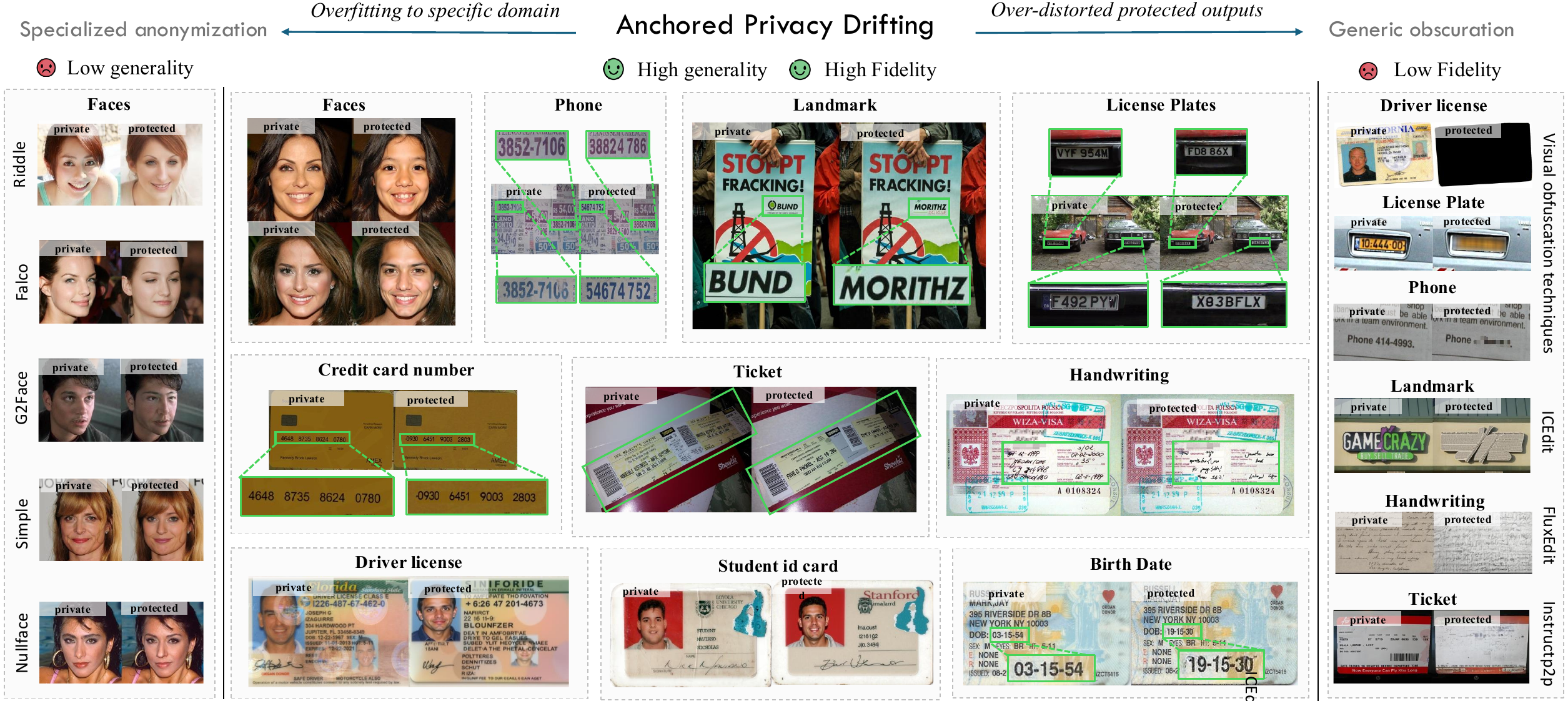}
    \caption{\textbf{Anchored Privacy Drifting (APD)} defines a new privacy protection paradigm for MLLMs. It preserves content coherence and high perceptual quality while explicitly maintaining information beneficial for \textit{downstream tasks} and \textit{personalized} user requirements across diverse inputs.}
    \label{fig:teaser}
    \vspace{-15pt}
\end{figure}
% \begin{center}
% \includegraphics[width=\textwidth]{figs/teaser.pdf}
% \captionof{figure}{}
% \label{fig:teaser}
% \end{center}
% To this end, we propose \textit{Privacy-Informed Editing (PIE)}, a training-free framework that generates a safe version of the source image with transformed private content, while preserving the overall visual fidelity, as illustrated in Fig~\ref{fig:teaser}. PIE operates in a shared multimodal latent space and steers the denoising trajectory with two directional signals to control where the generated image ultimately lands. 
% A high-level semantic direction leverages inherent semantic stochasticity to guide the trajectory toward a region semantically consistent with the source but disjoint from its private specifics. 
% A source-derived direction, computed directly between the source and the current state, anchors the trajectory against excessive deviation from the original context. 
% By adaptively balancing these two directions across spatial regions and generation steps, PIE controls the trajectory toward an endpoint that is structurally faithful to the source yet semantically redirected away from private specifics, achieving controllable privacy concealment with minimal contextual degradation.

 Given that current assessment protocols remain confined to task-specific indicators~\cite{riddle,falco,face_simple,g2face,nullface} (\textit{e.g.}, face embeddings) or subjective human evaluations~\cite{dipa2,human1}, we further present \textit{AdaptShield}, a benchmark encompassing 22 types of sensitive content with specified privacy regions and hierarchical evaluation protocols covering both concealment effectiveness and perceptual fidelity. For concealment, AdaptShield employs identity similarity, textual matching, and MLLM probing. For fidelity, it combines perceptual metrics, detection checks, and MLLM scoring. The two are further unified into a F1-Privacy metric to provide a holistic assessment of overall protection and fidelity performance. To mitigate model-specific bias, we employ four representative MLLMs as evaluators (Qwen2.5-VL-7B-Instruct~\cite{bai2025qwen25vltechnicalreport}, Qwen3-VL-8B-Instruct~\cite{qwen3}, InternVL3-14B~\cite{internvl3}, and InternVL3.5-14B~\cite{internvl35}). Extensive experiments demonstrate that APD achieves state-of-the-art performance across all 22 categories and four MLLM evaluators. 
 In summary, our contributions are as follows:
\begin{itemize}
\item We introduce \textit{Anchored Privacy Drifting (APD)}, a training-free, scalable framework that protects privacy by drifting sensitive regions toward privacy-safe alternatives while anchoring them to the source for visual fidelity, enabling adaptive sanitization of sensitive attributes while preserving contextual coherence.
\item We construct \textit{AdaptShield}, a comprehensive benchmark covering 22 heterogeneous privacy types with user-specified privacy masks and hierarchical evaluation protocols that jointly assess concealment effectiveness and perceptual fidelity, providing systematic and fine-grained evaluation for privacy protection in MLLM scenarios.
\item We conduct extensive experimental analyses demonstrating that APD consistently achieves substantial gains across diverse privacy categories, including an average 10.4\% improvement on textual privacy categories and an average 8.5\% improvement under MLLM-based evaluation across four MLLM evaluators.
\end{itemize}
\definecolor{myDeepG}{RGB}{6, 176, 80}
\begin{figure*}[t]
    \centering
    \includegraphics[width=0.95\linewidth]{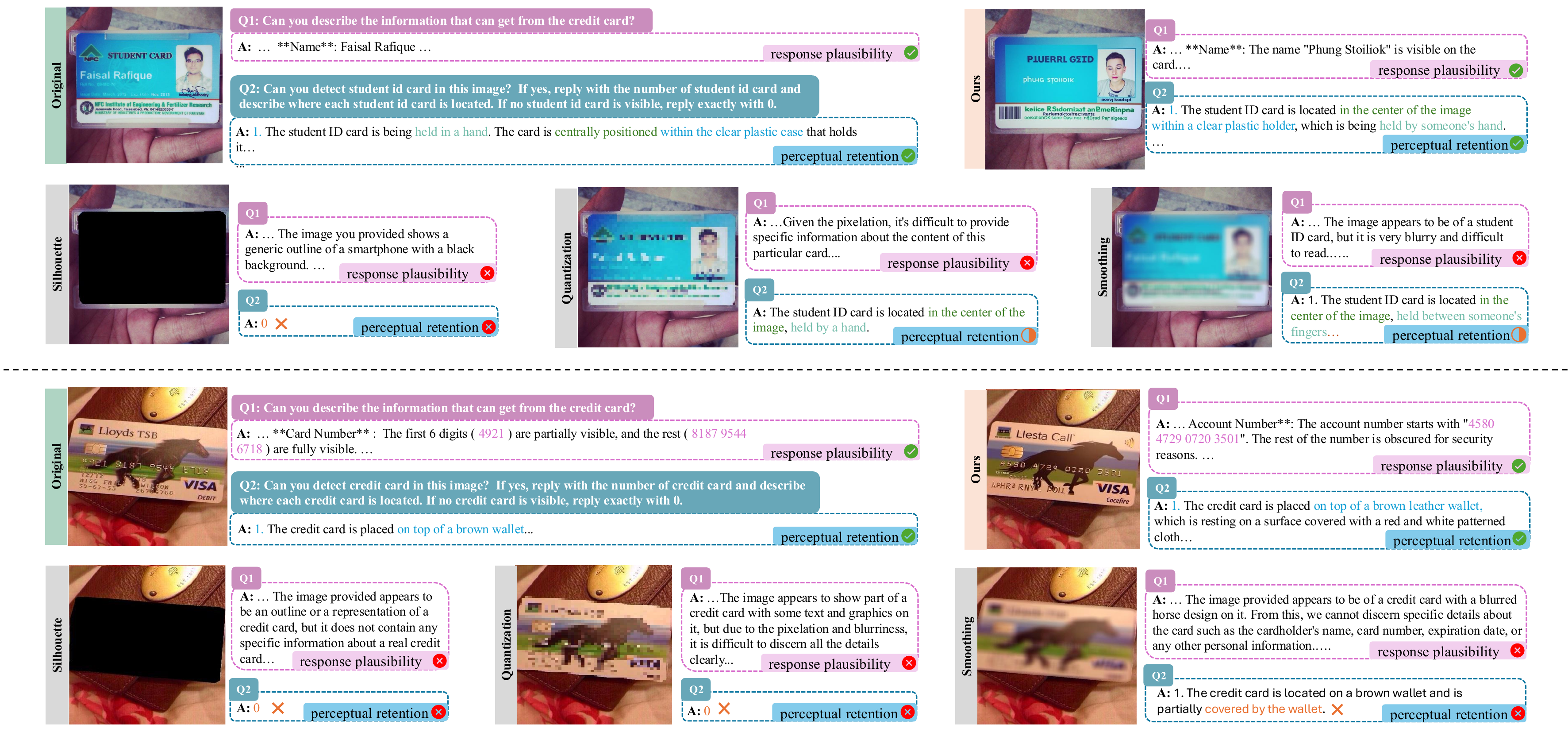}
    % \caption{Enter Caption}
    % \small
    \caption{
     Examples illustrating the necessity of high-fidelity visual privacy protection in MLLM scenarios. Simple obfuscation of sensitive regions impairs MLLM understanding, resulting in implausible responses (Q1) and degraded perceptual retention (Q2). In contrast, the proposed method preserves contextual cues, producing reasonable responses while safeguarding sensitive information. 
}
    \label{fig:extra}
    \vspace{-10pt}
\end{figure*}
\section{AdaptShield Benchmark}
\subsection{Data and Scope}

\begin{wrapfigure}[10]{r}{0.56\linewidth}
\vspace{-40pt}
\centering
\includegraphics[width=\linewidth]{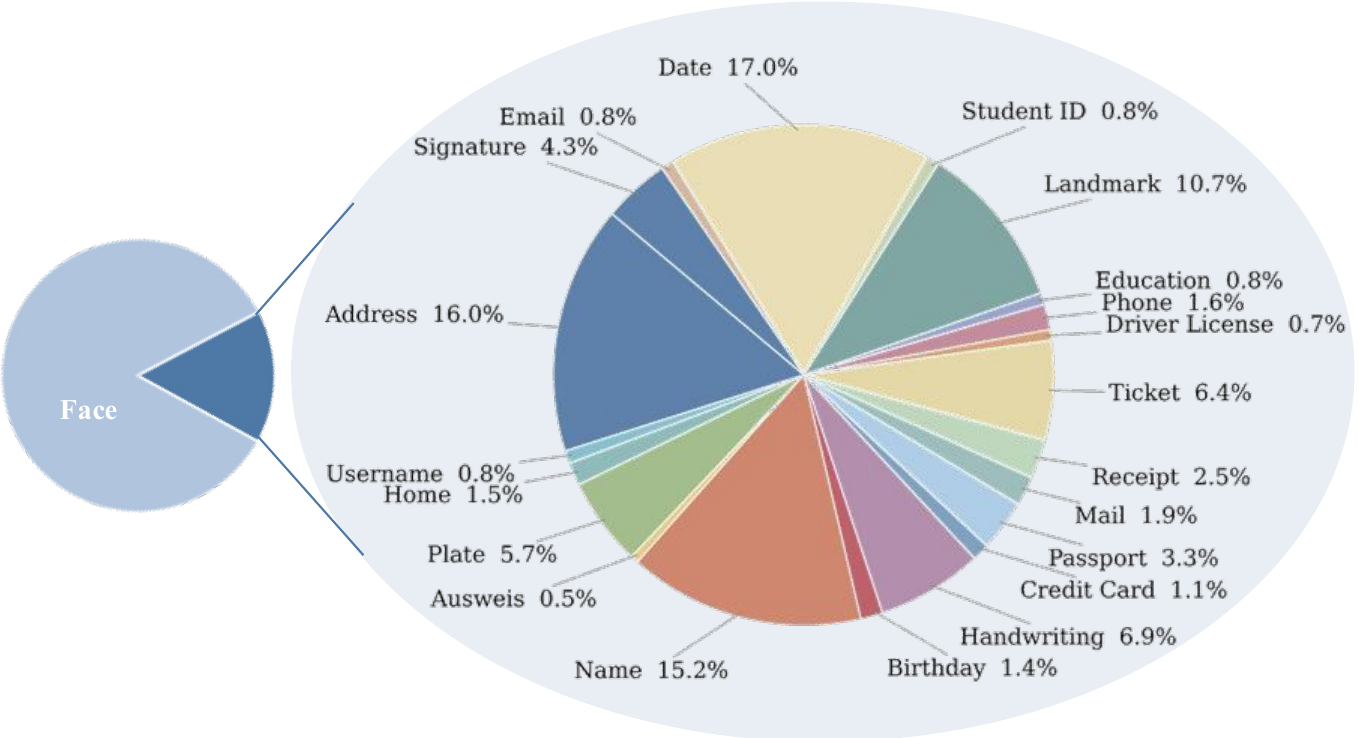}
% \vspace{-8pt}
\caption{Distribution of privacy types in AdaptShield.}
% \vspace{-8pt}
\label{fig:distribution}
\end{wrapfigure}
AdaptShield contains 32,491 images covering 22 privacy types, as shown in Fig.~\ref{fig:distribution}. 
For tailored analysis, we further organize these types into three groups:

\textbf{Face privacy} comprises 30,000 profile images that contain sensitive information about identity and appearance, which is the most studied private category.

\textbf{Textual privacy} contains 1,362 images across 12 symbolic or printed privacy categories,  including name (333), email (22), home address (37), mail (58), phone (24), birth date (22), username (19), address (326), receipt (69), date (367), credit card (41), and license plate (44).

\begin{wraptable}[10]{r}{0.39\linewidth}
\vspace{-12pt}
\centering
\small
\setlength{\tabcolsep}{2pt}
\caption{
Comparison between recent MLLM privacy benchmarks and our proposed \textbf{AdaptShield}.
}
\vspace{-6pt}
\resizebox{\linewidth}{!}{
\begin{tabular}{l|cc|cc}
\toprule
\multirow{2}{*}[-3pt]{\textbf{Benchmark}} & 
\multicolumn{2}{c|}{\textbf{Scale}} & 
\multicolumn{2}{c}{\textbf{Objective}} \\ 
\cmidrule(lr){2-3} \cmidrule(lr){4-5}
 & \textbf{Img.} & \textbf{Type}
 & \textbf{Leak.}  & \textbf{Qual.} 
\\ 
\midrule
CONFAIDE~\cite{CONFAIDE} & 766 & 10 &\textcolor{myDeepG}{\ding{51}} & \ding{55} \\
VIP~\cite{vip}& 554 & 8 &\textcolor{myDeepG}{\ding{51}} & \ding{55} \\
MLLMU~\cite{liu2025mllmu}  & 1,153& 1   &\textcolor{myDeepG}{\ding{51}} & \ding{55}  \\
Safe-LLaVA~\cite{safe_llava_2025} & 2,200& 1 & \textcolor{myDeepG}{\ding{51}} & \ding{55} \\
MLLMGuard~\cite{gu2024mllmguardmultidimensionalsafetyevaluation}& 323 & - & \textcolor{myDeepG}{\ding{51}}  & \ding{55}\\
Multitrust~\cite{zhang2024multitrust} & 1,300& 16 &\textcolor{myDeepG}{\ding{51}}  & \ding{55}\\
PII-Bench~\cite{pii} & 4,000 & 20 &\textcolor{myDeepG}{\ding{51}}  & \ding{55} \\
OutSafe~\cite{outsafe} & 23,400 & 9&\textcolor{myDeepG}{\ding{51}}  & \ding{55} \\
\rowcolor{blue!5}
\textbf{AdaptShield} & 32,491 & 22 & \textcolor{myDeepG}{\ding{51}} &\textcolor{myDeepG}{\ding{51}}  \\
\bottomrule
\end{tabular}
}
\vspace{-50pt}
\label{tab:privacy-benchmark-comparison}
\end{wraptable}
\textbf{Composite privacy} includes 1,129 images of 9 types with intertwined visual and textual privacy signals, such as education (28), passport (124), ausweis (19), student identification (31), driver license (25), handwriting (255), ticket (240), signature (160) and landmark (247). 

Raw images are collected from  public databases CelebAMask-HQ~\cite{CelebAMask-HQ} and VISPR~\cite{vispr}, avoiding the introduction of new privacy risks. Category-specific region annotations are constructed from the original dataset annotations and manually verified.
% \noindent\textbf{Scope.}  Our benchmark covers 22 types of privacy, which are categorized according to their representation forms. 
% (1) Face privacy (1 type) is the most extensively studied category and typically has domain-specific metrics or models for evaluation. 
% (2) Textual privacy (12 types) involves symbolic or printed content such as license plates, identification   , or credit card digits, which can be extracted as textual information. 
% (3) Composite privacy (9 types) refers to cases where visual and textual cues either coexist within the same scene (\textit{e.g.,} a document containing both a profile and text) or are intertwined within a single modality (\textit{e.g.,} handwritten notes), which require more integrated reasoning for accurate evaluation.

\subsection{Benchmark Objectives}
\label{metric}

As shown in Tab.~\ref{tab:privacy-benchmark-comparison}, existing benchmarks mainly evaluate privacy leakage (\textit{Leak.}) while overlooking the quality and usability (\textit{Qual.}) of protected images.
Fig.~\ref{fig:extra} shows that low-fidelity protection may remove critical contextual cues, leading to implausible MLLM responses or degraded visual retention.
This indicates that visual quality is essential for effective privacy protection in MLLM scenarios.
To this end, the proposed AdaptShield aims to jointly evaluate privacy concealment and content fidelity under open-domain and personalized MLLM settings. \vspace{2pt}
\\
\noindent 
\noindent\textbf{Protection score $\boldsymbol{P}$.}  
We assess privacy leakage using evaluation protocols tailored to different privacy types. For face privacy, identity similarity is measured using a pre-trained recognition encoder. For textual privacy, OCR-extracted content is compared against the original sensitive text. For composite privacy, leakage-oriented MLLM probing is used to assess whether sensitive information remains recognizable.

% 这里可以加一下可跳转的ref～
% \noindent\textbf{Protection score $\boldsymbol{P}$.}  
% We apply type-specific metrics tailored to different privacy modalities. 
% (1) For identity-related privacy (\textit{e.g.,} faces), we measure feature similarity using a pre-trained recognition encoder to quantify identity leakage. 
% (2) For textual or symbolic privacy, we extract content via Optical Character Recognition (OCR) and compare string-level overlap between the protected and original content. 
% (3) For complex or composite privacy types, we employ leakage-oriented probing: MLLMs are asked to describe sensitive information in the image, and the generated descriptions are compared with the ground-truth annotations to determine which sensitive elements remain recognizable.
% \smallskip
\noindent\textbf{Fidelity score $\boldsymbol{F}$.}  
We evaluate fidelity separately for each privacy category. For face privacy, SSIM, PSNR, and face detection consistency are used to measure structural preservation. For textual privacy, OCR readability is adopted to assess whether the protected content remains visually coherent and readable. For composite privacy, MLLM-based evaluation is adopted to assess semantic coherence and visual quality.

% % 这里可以加一下可跳转的ref～
% \noindent\textbf{Fidelity score $\boldsymbol{F}$.}  
% To assess perceptual quality and contextual integrity of protected images, we also employ modality-specific fidelity metrics. 
% (1) For faces, we adopt SSIM and PSNR to measure preservation of non-sensitive information, and use a face detector to assess whether the protected output still preserves valid facial structure.  (2) For text or symbols, we evaluate whether OCR systems can still extract coherent and legible content without distortion. (3) For complex or composite scenes, we apply MLLM-based scoring to assess semantic coherence, image quality, and overall visual realism. \\

\textbf{Overall Score F1-privacy.}  
We compute an overall score by combining Protection ($P$) and Fidelity ($F$) using the harmonic mean~\cite{f1_score,10.1145/3606367}. 
This comprehensive score, denoted as $\text{F1-privacy}$, is defined as
$\text{F1-privacy} = (2 \cdot P \cdot F)/(P+F)$.
All metrics are normalized to $[0,1]$, and lower-is-better metrics are inverted so that higher values consistently indicate better protection or fidelity.
% All Protection ($P$) and Fidelity ($F$) scores are normalized to the range $[0,1]$. 
% Let $\mathcal{S}$ denote the set of protection metrics. For protection metrics $s$ where lower values indicate better privacy, we first invert them using $1 - s$ so that higher values consistently reflect stronger protection. 

% The resulting F1-privacy score combines $P$ and $F$ while penalizing extreme imbalances, such as excessive obfuscation that degrades perceptual quality or high fidelity that risks leaking private information, yielding a fair and interpretable overall metric.

\section{Anchored Privacy Drifting (APD)}
\begin{figure*}[h]
    \centering
    % \vspace{1mm}
    \includegraphics[width=1\linewidth]{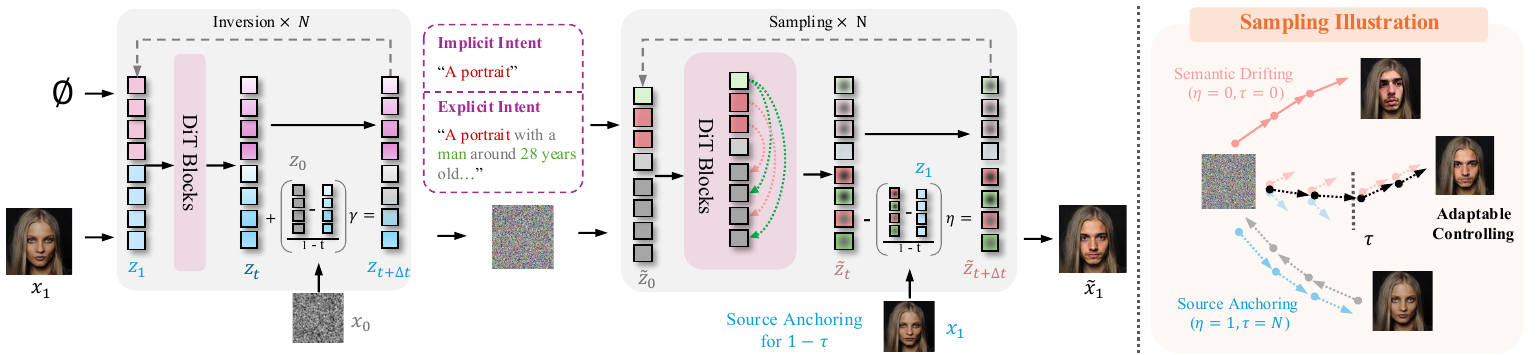}
\caption{\textbf{\textit{Left.}} Overview of the proposed Anchored Privacy Drifting (APD) framework.
The source image is inverted to a latent noise space and regenerated under adaptive guidance from semantic drifting and source anchoring.
\textbf{\textit{Right.}}  Compared to destructive protection (top) or rigid preservation (bottom), our adaptive controller (middle) achieves privacy-preserving yet faithful outputs.}
    \label{fig:structure}
    \vspace{-8pt}
\end{figure*}
\subsection{Overview}
Privacy protection for MLLM scenarios poses an asymmetric controllable generation problem where sensitive attributes must be transformed while contextual content stays close to the source. We address this by formulating the task as a trajectory control problem in a shared multimodal latent space. As illustrated in the right panel of Fig.~\ref{fig:structure}, this steering is realized through two complementary vector fields: a semantic drift field that drives sensitive attributes away from the source, and a source anchoring field that regularizes the trajectory against contextual deviation. Dynamically composing these two fields yields a trajectory toward a structurally faithful yet privacy-safe endpoint.

\subsection{Privacy-aware Semantic Drifting}
% We build our framework upon a Multimodal Diffusion Transformer (MM-DiT) structure~\cite{mmdit}, where image latents and text semantics are jointly embedded and interact within a shared multimodal latent space. Within this shared space, encoded semantic conditions enable the estimation of a semantic vector field that transports the image-latent trajectory toward the desired semantic region.

Our framework is build upon a Multimodal Diffusion Transformer (MMDiT)~\cite{flux}, where image latents and text semantics are jointly embedded into a shared multimodal latent space. Producing the privacy-safe image is modeled as progressively transporting an initial image latent along a trajectory toward a target endpoint, which is decoded into the final result. The evolution of the trajectory at each step is determined by a directional vector field, and steering this field accordingly allows us to control where the endpoint ultimately lands.
In our setting, this control is realized through a textual semantic condition $c$.  Leveraging the joint image–text embedding, the model predicts a text-conditioned drift field $v_{\phi}$ that points the latent state toward the semantic region specified by $c$:
\begin{equation}
v_{\phi}(z_t,t,c),
\end{equation}
where $z_t$ denotes the latent state at timestep $t$. At each denoising step, following this vector field progressively transports the latent trajectory toward a region matching the semantic description.

This formulation supports flexible semantic guidance. The user can explicitly specify target attributes (e.g., redirecting "female" to "male"), deterministically driving those attributes toward the corresponding semantic manifold. Any unspecified attributes remain free, and are filled in by the model's stochastic sampling, generating plausible details that naturally diverge from the original sensitive attributes. Crucially, this stochastic divergence is itself a privacy-preserving mechanism: even without explicit intent, the trajectory still drifts away from the original private specifics.

% The resulting vector field defines the instantaneous transport direction at each denoising timestep, progressively steering the latent trajectory toward a privacy-safe semantic manifold.
% For example, with private information change requirement, the trajectory can be explicitly guided toward another semantic concept such as ``female'', thereby moving away from the original privacy-sensitive identity manifold. Importantly, explicit attribute replacement is not necessary for privacy transformation.
% Even high-level semantic conditions, such as ``a person'', introduce stochastic semantic transport during generation.
% Since the diffusion trajectory inherently contains randomness, the generated identity naturally deviates from the original sensitive attributes while remaining semantically plausible.
% This property allows the framework to support both personalized privacy editing with explicit user intent and generic privacy protection through naturally induced semantic ambiguity.

\subsection{Source Anchoring Path}

The semantic drifting defined above steers the trajectory toward privacy-safe semantics, achieving privacy transformation.  However, stochasticity that diversifies private details also makes the trajectory drift uncontrollably, leading to excessive modifications that unintentionally degrade contextual content. To preserve contextual fidelity, we introduce a source anchoring field that continuously constrains the trajectory toward the source image manifold. In contrast to the semantic drift field, which pushes the trajectory away from the original privacy-sensitive region, the source anchoring field counteracts excessive drift by pulling the trajectory back toward the source.

Let $z_1$ denote the latent representation of the source image. The initialization latent $\tilde{z}_0$ is obtained by injecting random noise $z_0$ into $z_1$. Starting from $\tilde{z}_0$, the trajectory is then guided by a vector pointing from the current state $z_t$ and $z_1$ at each step, yielding the source anchoring velocity field:
\begin{equation}
u(z_t|z_1)=\frac{z_1-z_t}{1-t},
\end{equation}
where $u(z_t|z_1)$  denotes the source-conditioned velocity that transports $z_t$ along the straight path toward $z_1$, derived from the linear-interpolation assumption $z_t=(1-t)z_0+tz_1$ introduced in ~\cite{rf}. Exclusively integrating this field at each step transports $\tilde{z}_0$ back to $z_1$. As its strength varies, the trajectory's binding to the source manifold scales accordingly, enabling tunable source preservation.
% Exclusively integrating this field at each step reconstructs the source image. As its strength varies, the trajectory's binding to the source manifold scales accordingly.

\subsection{Adaptive Vector Field Composition}

Since both vector fields provide transport directions at each denoising timestep within the same latent space, the final generation trajectory can be formulated as a dynamic composition of the semantic drift field and the source anchoring field. Specifically, the final transport field is defined as:
\begin{equation}
\tilde{v}(z_t,t)
=
v_{\phi}(z_t,t,c)
+
\eta_t(p)\,u(z_t|z_1),
\end{equation}
where $\eta_t(p)$ controls the anchoring strength at timestep $t$ and spatial position $p$.
Under this formulation, the semantic drift field enables privacy transformation, while the source anchoring field maintains contextual fidelity. The final trajectory is therefore shaped by their interplay across spatial regions and denoising steps, jointly determining where the generated image ultimately lands. This control is achieved by the spatially and temporally adaptive coefficient $\eta_t(p)$:
\begin{equation}
\eta_t(p)
=
\mathbf{1}[p \notin \mathcal{M}]
+
\eta_0 \cdot \mathbf{1}[p \in \mathcal{M},\, t < \tau],
\end{equation}
where $\mathcal{M}$ denotes the sensitive spatial region, $\tau$ is a timestep threshold, and $\eta_0 \in (0,1)$ controls partial anchoring strength.
This formulation produces different transport behaviors across regions and timesteps.
For non-sensitive regions, the anchoring field dominates throughout generation, enforcing strong structural consistency with the source. For sensitive regions, anchoring stays active in early steps to preserve coarse layout, then weakens after timestep $\tau$ to let the drift field take over and complete the semantic transformation. By modulating the balance between drift and anchoring across regions and stages, the trajectory diverges from privacy-sensitive content while staying close to the overall source context, yielding privacy-safe outputs with preserved contextual fidelity.

\section{Experiments}

\begin{figure*}[t]
    \centering
    \includegraphics[width=1\linewidth]{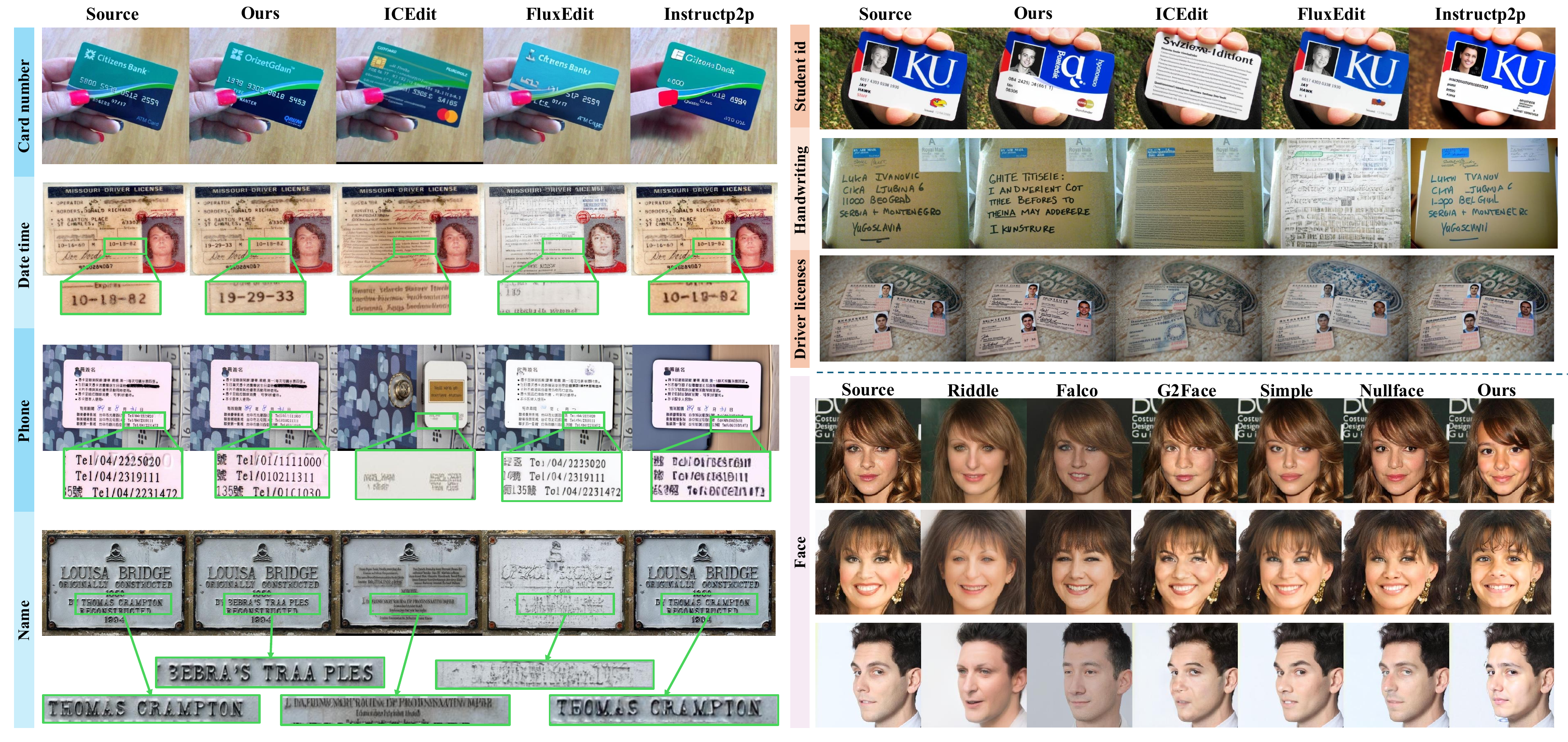}
    \caption{
    \setlength{\fboxsep}{1pt}
% Qualitative Visualization of Textual, Composite, and Facial Privacy Categories. Category labels for \colorbox{cyan!15}{textual\vphantom{Ty}}, \colorbox{orange!15}{composite\vphantom{Ty}}, and \colorbox{pink!30}{facial\vphantom{Ty}} privacy are shown with \colorbox{cyan!15}{blue\vphantom{Ty}}, \colorbox{orange!15}{orange\vphantom{Ty}}, and \colorbox{pink!30}{pink\vphantom{Ty}} background colors for clearer distinction, respectively. Our method not only obfuscates sensitive details but also maintains stylistic and semantic alignment with the original content, resulting in coherent and natural outputs after protection. In particular, for \colorbox{pink!30}{Facial\vphantom{Ty}} regions, unlike prior fixed anonymization methods, our approach allows adaptive adjustment of sensitive attributes such as age, ethnicity, and gender, achieving more personalized and versatile privacy-preserving results.
Qualitative Visualization of Textual, Composite, and Facial Privacy Categories.
Category labels are color-coded as \colorbox{cyan!15}{textual\vphantom{Ty}}, \colorbox{orange!15}{composite\vphantom{Ty}}, and \colorbox{pink!30}{facial\vphantom{Ty}} for clearer distinction.
Our method not only obfuscates sensitive details but also maintains stylistic and semantic alignment with the original content, resulting in coherent and natural outputs after protection. In particular, for \colorbox{pink!30}{facial\vphantom{Ty}} regions, unlike prior fixed anonymization methods, our approach enables adaptive adjustment of sensitive attributes such as \textit{age}, \textit{ethnicity}, and \textit{gender}, achieving more personalized and versatile privacy-preserving results. Please zoom in for better visualization.
}
\vspace{-10pt}
\label{fig:compares}
\end{figure*}
% \smallskip
\noindent \textbf{Datasets.} 
All experiments are conducted on our AdaptShield benchmark, which contains 32,491 images spanning 22 privacy categories.
AdaptShield is constructed from publicly available datasets, including CelebAMask-HQ~\cite{CelebAMask-HQ} and VISPR~\cite{vispr}, and does not introduce newly collected private or personally identifiable data. For face anonymization, we use samples derived from CelebAMask-HQ to ensure a fair comparison with existing face anonymization baselines.

\smallskip
\noindent \textbf{Baselines.}
For face anonymization, we compare with representative methods including RIDDLE~\cite{riddle} and FALCO~\cite{falco}, which emphasize strong privacy protection, Simple~\cite{face_simple} and G2Face~\cite{g2face}, which rely on face swapping to achieve visually consistent results, and NullFace~\cite{nullface}, a training-free method similar to our approach. For textual and composite privacy types, we adopt standard obscuration-based baselines, including Silhouette~\cite{li2017effectiveness}, Smoothing~\cite{hill2016effectiveness}, and Quantization~\cite{fan2018image}. We further compare with representative instruction-conditioned visual transformation models, including InstructPix2Pix~\cite{instructp2p}, FluxEdit~\cite{fluxedit}, and ICEdit~\cite{icedit}. These models can alter image content according to textual instructions, offering a general comparison for controllable sensitive-content transformation.

\noindent \textbf{Metrics.}
For all categories, we compute F1-privacy score as described in Section ~\ref{metric}, which quantifies the balance between privacy protection and visual fidelity. The definitions of Protection score ($P$) and Fidelity score ($F$) are tailored to each privacy type. In face anonymization, $P$ is evaluated using cosine similarity and $L_2$ distance between original and protected embeddings extracted by VGG-Face, FaceNet, and ArcFace backbones via the DeepFace framework~\cite{deepface}. Score $F$ is assessed on non-sensitive regions using SSIM and PSNR and face detection rate~\cite{deepface} for facial preservation. For textual regions, $F$ is measured using the OCR-detection rate. The protection score $P$ is computed as the mismatch between the extracted text and the original sensitive content. For composite categories, and for complex textual categories where OCR-based extraction may be unreliable, we use MLLM evaluators, where $F$ is measured as the average MLLM assessment score across plausibility, consistency, and visual quality, and $P$ is calculated by averaging BLEU, ROUGE, and METEOR scores between MLLM-generated descriptions of the sensitive content and the original.

\textbf{MLLM Evaluator.} We select four widely used MLLMs for evaluation to avoid model bias: Qwen3-VL-8B-Instruct~\cite{qwen3}, Qwen2.5-VL-7B-Instruct~\cite{bai2025qwen25vltechnicalreport}, InternVL3-14B~\cite{internvl3}, and InternVL3.5-14B~\cite{internvl35}. These models are chosen to cover diverse architectures and scales, reducing potential model bias in our evaluation.

% \subsection{Main results}
% \paragraph{Comparison with face anonymization baselines.}
% \paragraph{Face anonymization} 

% \begin{figure*}[htbp]
%     \centering
%     \includegraphics[width=1\linewidth]{sec/Figs/res_face.pdf}
%     \caption{Examples of controllable face anonymization. Our method flexibly adjusts age, gender, and ethnicity while maintaining natural and coherent appearances.}
%     \label{fig:face_viz}
% \end{figure*}
% \vspace{2mm}
\subsection{Main results}
In this section, we comprehensively evaluate our method in terms of its ability to jointly balance privacy protection ($P$) and visual fidelity ($F$). To provide a holistic assessment, we combine conventional quantitative metrics, MLLM-based evaluation, and qualitative comparisons to demonstrate both the effectiveness and perceptual quality of our approach. Our evaluation is organized into three tracks: (1) Textual obfuscation, (2) MLLM-based evaluation, and (3) Face anonymization.

\noindent \textbf{Textual Obfuscation.} As illustrated in Tab.~\ref{tab:text}, 
% existing textual obscuration baselines either preserve text readability with low-intensity protection, resulting in a low protection strength and privacy leakage, or increase protection strength with high-intensity protection at the cost of perceptual distortion. In contrast, o
our method achieves state-of-the-art performance, with a 10.4\% improvement over the best baseline, FluxEdit~\cite{fluxedit}. Besides, although methods like Silhouette and Smoothing can provide strong protection, their ability to preserve content is extremely limited, resulting in very low F1-privacy scores. Unlike these destructive approaches, our approach maintains a readable context while effectively altering the original content. This is further evidenced by the qualitative examples in Fig.~\ref{fig:compares}, where a credit card number “5800 5978 0512 2559” is transformed into “1939 9302 8618 5453,” preserving the 4-digit grouping while avoiding any overlap with the original digits. Both qualitative and quantitative results show that our method preserves structure and semantics while effectively protecting sensitive information. Although OCR can be unstable under complex layouts, its results remain useful references for surface-level textual leakage and repetition. 
% Meanwhile, our MLLM-based evaluation track also covers complex textual privacy cases and provides complementary evidence that corroborates the effectiveness observed in the textual track.
\begin{table}[t]
\centering
% \vspace{-10pt}
\setlength{\tabcolsep}{4pt}
\caption{Performance comparison of the F1-privacy (P/F) score based on multiple representative MLLMs, including \textit{Qwen3VL}, \textit{Qwen2.5VL}, \textit{InternVL3}, and \textit{InternVL3.5}. Fidelity ($F$) is computed as the average of three normalized MLLM-evaluated scores: Semantic Plausibility, Perceptual Consistency, and Visual Quality. 
Protection ($P$) is defined as $P = 1 - \text{Similarity}$, where Similarity is calculated as the average of BLEU, ROUGE-1, and METEOR scores between the MLLM description of sensitive content.
% Protection ($P$) is calculated as $1$ minus the average Similarity, where Similarity is measured using BLEU, ROUGE-1, and METEOR between the MLLM-generated description of sensitive content.
% Our method achieves the highest harmonic mean, demonstrating a superior trade-off between privacy protection and output fidelity. 
 We report the aggregated results as mean $\pm$ standard deviation.}
\label{tab:privacy_results_main}
% \vspace{-2mm}
\resizebox{\textwidth}{!}{%
\begin{tabular}{l|ccccccccccc|c}
\toprule
\textbf{Method} & \textbf{Education} & \textbf{Signature} & \textbf{Ticket} & \textbf{Handwriting} & \textbf{Student ID} & \textbf{Mail} & \textbf{Driver License} & \textbf{Ausweis} & \textbf{Credit Card} & \textbf{Receipt} & \textbf{Landmark} & \textbf{Avg} \\
\midrule
\rowcolor{blue!5}
\textbf{APD} & \textcolor{red}{{\normalsize 0.840}{\scriptsize $\pm$0.078}} & \textcolor{red}{{\normalsize 0.820}{\scriptsize $\pm$0.051}} & \textcolor{red}{{\normalsize 0.794}{\scriptsize $\pm$0.026}} & \textcolor{red}{{\normalsize 0.818}{\scriptsize $\pm$0.025}} & \textcolor{red}{{\normalsize 0.802}{\scriptsize $\pm$0.029}} & \textcolor{red}{{\normalsize 0.819}{\scriptsize $\pm$0.026}} & \textcolor{red}{{\normalsize 0.774}{\scriptsize $\pm$0.028}} & \textcolor{red}{{\normalsize 0.781}{\scriptsize $\pm$0.012}} & \textcolor{red}{{\normalsize 0.804}{\scriptsize $\pm$0.031}} & \textcolor{red}{{\normalsize 0.762}{\scriptsize $\pm$0.031}} & \textcolor{red}{{\normalsize 0.858}{\scriptsize $\pm$0.062}} & \textcolor{red}{{\normalsize 0.821}{\scriptsize $\pm$0.035}} \\
\textbf{ICEdit} & \textcolor{blue}{{\normalsize 0.781}{\scriptsize $\pm$0.078}} & {\normalsize 0.778}{\scriptsize $\pm$0.046} & {\normalsize 0.755}{\scriptsize $\pm$0.025} & {\normalsize 0.787}{\scriptsize $\pm$0.049} & \textcolor{blue}{{\normalsize 0.732}{\scriptsize $\pm$0.017}} & \textcolor{blue}{{\normalsize 0.747}{\scriptsize $\pm$0.034}} & {\normalsize 0.680}{\scriptsize $\pm$0.025} & {\normalsize 0.731}{\scriptsize $\pm$0.014} & \textcolor{blue}{{\normalsize 0.749}{\scriptsize $\pm$0.034}} & {\normalsize 0.721}{\scriptsize $\pm$0.038} & {\normalsize 0.746}{\scriptsize $\pm$0.029} & \textcolor{blue}{{\normalsize 0.757}{\scriptsize $\pm$0.034}} \\
\textbf{InstructP2P} & {\normalsize 0.777}{\scriptsize $\pm$0.075} & {\normalsize 0.779}{\scriptsize $\pm$0.046} & \textcolor{blue}{{\normalsize 0.762}{\scriptsize $\pm$0.026}} & \textcolor{blue}{{\normalsize 0.791}{\scriptsize $\pm$0.035}} & {\normalsize 0.719}{\scriptsize $\pm$0.033} & {\normalsize 0.742}{\scriptsize $\pm$0.041} & \textcolor{blue}{{\normalsize 0.738}{\scriptsize $\pm$0.044}} & \textcolor{blue}{{\normalsize 0.740}{\scriptsize $\pm$0.035}} & {\normalsize 0.744}{\scriptsize $\pm$0.044} & \textcolor{blue}{{\normalsize 0.752}{\scriptsize $\pm$0.024}} & {\normalsize 0.707}{\scriptsize $\pm$0.027} & {\normalsize 0.749}{\scriptsize $\pm$0.031} \\
\textbf{FluxEdit} & {\normalsize 0.730}{\scriptsize $\pm$0.053} & {\normalsize 0.758}{\scriptsize $\pm$0.039} & {\normalsize 0.712}{\scriptsize $\pm$0.022} & {\normalsize 0.756}{\scriptsize $\pm$0.033} & {\normalsize 0.649}{\scriptsize $\pm$0.029} & {\normalsize 0.724}{\scriptsize $\pm$0.022} & {\normalsize 0.637}{\scriptsize $\pm$0.059} & {\normalsize 0.677}{\scriptsize $\pm$0.036} & {\normalsize 0.680}{\scriptsize $\pm$0.033} & {\normalsize 0.708}{\scriptsize $\pm$0.029} & {\normalsize 0.693}{\scriptsize $\pm$0.022} & {\normalsize 0.716}{\scriptsize $\pm$0.026} \\
\textbf{Silhouette} & {\normalsize 0.666}{\scriptsize $\pm$0.093} & {\normalsize 0.775}{\scriptsize $\pm$0.051} & {\normalsize 0.536}{\scriptsize $\pm$0.079} & {\normalsize 0.660}{\scriptsize $\pm$0.054} & {\normalsize 0.453}{\scriptsize $\pm$0.059} & {\normalsize 0.453}{\scriptsize $\pm$0.064} & {\normalsize 0.463}{\scriptsize $\pm$0.061} & {\normalsize 0.488}{\scriptsize $\pm$0.073} & {\normalsize 0.589}{\scriptsize $\pm$0.079} & {\normalsize 0.486}{\scriptsize $\pm$0.061} & \textcolor{blue}{{\normalsize 0.792}{\scriptsize $\pm$0.072}} & {\normalsize 0.657}{\scriptsize $\pm$0.057} \\
\textbf{Smoothing} & {\normalsize 0.727}{\scriptsize $\pm$0.089} & \textcolor{blue}{{\normalsize 0.786}{\scriptsize $\pm$0.059}} & {\normalsize 0.675}{\scriptsize $\pm$0.036} & {\normalsize 0.713}{\scriptsize $\pm$0.037} & {\normalsize 0.655}{\scriptsize $\pm$0.049} & {\normalsize 0.663}{\scriptsize $\pm$0.049} & {\normalsize 0.619}{\scriptsize $\pm$0.035} & {\normalsize 0.625}{\scriptsize $\pm$0.054} & {\normalsize 0.684}{\scriptsize $\pm$0.055} & {\normalsize 0.657}{\scriptsize $\pm$0.022} & {\normalsize 0.709}{\scriptsize $\pm$0.064} & {\normalsize 0.703}{\scriptsize $\pm$0.043} \\
\textbf{Quantization} & {\normalsize 0.692}{\scriptsize $\pm$0.048} & {\normalsize 0.759}{\scriptsize $\pm$0.058} & {\normalsize 0.629}{\scriptsize $\pm$0.014} & {\normalsize 0.681}{\scriptsize $\pm$0.041} & {\normalsize 0.623}{\scriptsize $\pm$0.018} & {\normalsize 0.643}{\scriptsize $\pm$0.042} & {\normalsize 0.612}{\scriptsize $\pm$0.041} & {\normalsize 0.623}{\scriptsize $\pm$0.019} & {\normalsize 0.635}{\scriptsize $\pm$0.014} & {\normalsize 0.642}{\scriptsize $\pm$0.033} & {\normalsize 0.696}{\scriptsize $\pm$0.057} & {\normalsize 0.677}{\scriptsize $\pm$0.031} \\
\bottomrule
\end{tabular}%
}
    \centering
    \includegraphics[width=1.\linewidth]{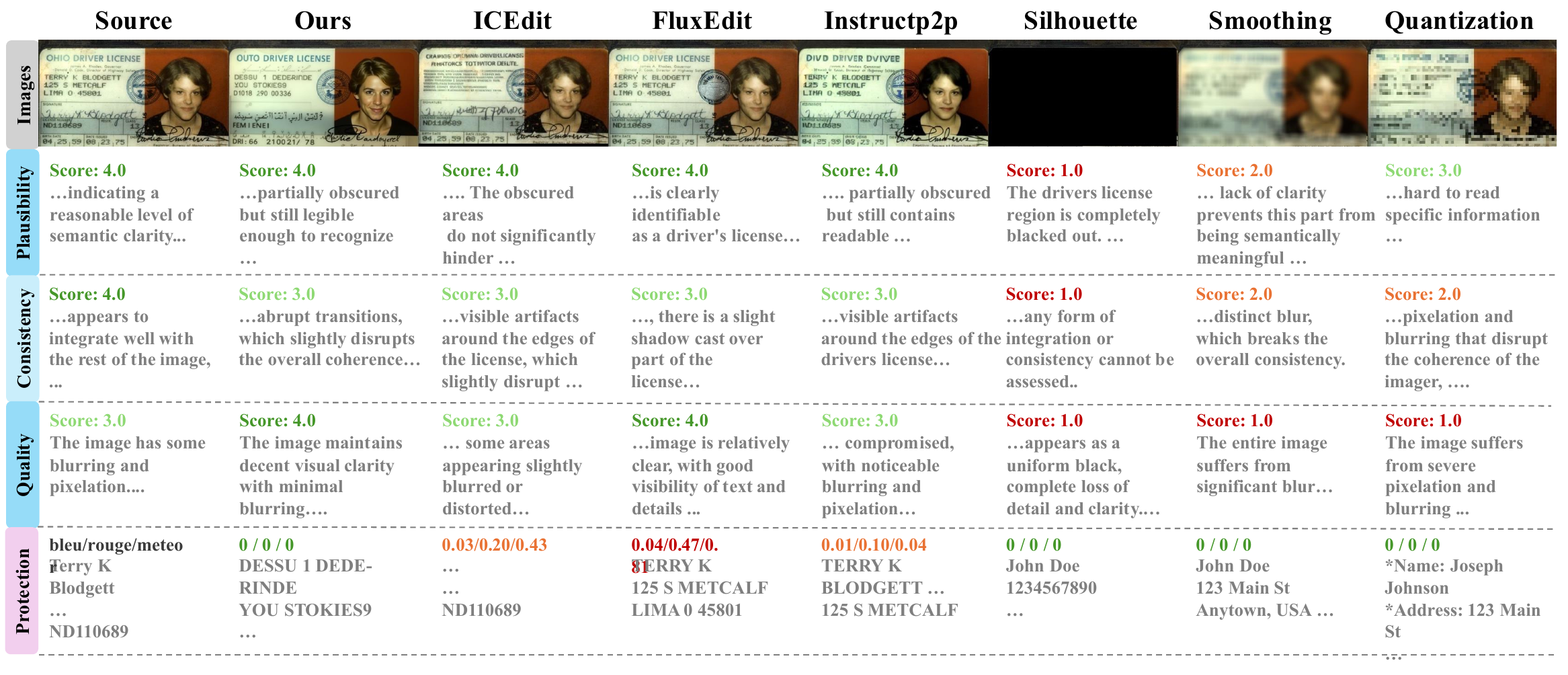}
    % \vspace{0.1mm}
\captionof{figure}{Illustration of MLLM-based assessment using Qwen2.5-VL-7B-Instruct. 
Given an input image, the MLLM directly outputs scores for \textit{semantic plausibility}, \textit{perceptual consistency}, and \textit{visual quality} to define the fidelity score $F$. For protection effectiveness $P$, the MLLM first generates a textual description of the sensitive content in the protected image, which is then compared against the original description via lexical similarity metrics (BLEU, ROUGE, and METEOR).}
\label{fig:mllm-example}
\vspace{-8pt}
\end{table}

\begin{table*}[t]
\centering
\small
\caption{Performance comparison of the {F1-privacy} score across textual categories. Fidelity $F$ is the Detection Ratio, representing structural preservation, and Protection $P$ is defined as $1 - \text{Hit Ratio}$, where the Hit Ratio measures text overlap and a higher $P$ indicates stronger privacy protection.}
\setlength{\tabcolsep}{4pt}
\resizebox{\textwidth}{!}{\begin{tabular}{l|cccccccccccc|c}
\toprule
\textbf{{Method}} & \textbf{Phone} & \textbf{Username} & \textbf{Date} & \textbf{Receipt} & \textbf{Mail} & \textbf{ Home} & \textbf{Address} & \textbf{Birthday} & \textbf{CreditCard} & \textbf{Plate} & \textbf{Email} & \textbf{Name}&\textbf{{Avg}} \\
\midrule
APD & \textbf{0.344} & \textbf{0.877} & \textbf{0.474} & \textbf{0.693} & \textbf{0.829} & \textbf{0.617} & \textbf{0.535} & \textbf{0.524} & \textbf{0.645} & \textbf{0.340} & \textbf{0.692} & \textbf{0.532}& \textbf{0.592}\\
FluxEdit~\cite{fluxedit} & 0.307 & 0.725 & 0.472 & \underline{0.688} & 0.812 & \underline{0.616} & \underline{0.534} & 0.419 & 0.378 & 0.332 & \underline{0.616} & \underline{0.530}& \underline{0.536}\\
InstructP2P~\cite{instructp2p} & 0.280 & \underline{0.794} & \underline{0.473} & 0.651 & \underline{0.821} & 0.512 & 0.420 & \underline{0.478} & \underline{0.472} & 0.238 & 0.502 & 0.499& 0.512 \\
ICEdit~\cite{icedit} & 0.207 & 0.525 & 0.263 & 0.464 & 0.752 & 0.347 & 0.305 & 0.000 & 0.216 & \textbf{0.340} & 0.361 & 0.322& 0.342 \\
Quantization~\cite{fan2018image} & \underline{0.343} & 0.348 & 0.315 & 0.462 & 0.729 & 0.351 & 0.437 & 0.000 & 0.179 & 0.205 & 0.239 & 0.385& 0.333 \\
Smoothing~\cite{hill2016effectiveness} & 0.000 & 0.272 & 0.137 & 0.207 & 0.410 & 0.102 & 0.099 & 0.000 & 0.134 & 0.000 & 0.086 & 0.102& 0.129 \\
Silhouette~\cite{li2017effectiveness} & 0.000 & 0.000 & 0.053 & 0.028 & 0.000 & 0.000 & 0.035 & 0.000 & 0.000 & 0.000 & 0.000 & 0.047& 0.014 \\
\bottomrule
\end{tabular}}
\label{tab:text}
\end{table*}

\begin{table*}[t]
    \centering
    \caption{Performance comparison on the F1-Privacy ($P$ / $F$) score for identity categories. Protection ($P$) is measured by the cosine 
similarity of identity features from VGGFace (VGG), FaceNet (FN), and 
ArcFace (AF). Fidelity ($F$) combines  perceptual metrics (SSIM and normalized PSNR, where PSNR 
is normalized by 40) and a structural integrity indicator 
(face detection rate). 
{Train-free} indicates whether the method requires no 
additional training, and {General} indicates 
whether the method generalizes beyond face-centric scenarios to broader 
privacy categories.}
% \vspace{-2mm}
\setlength{\tabcolsep}{2pt}
    \label{tab:face_tradeoff_scores}
    \resizebox{\textwidth}{!}{
    \begin{tabular}{l|cc|ccc|ccc|ccc|c}
        \toprule
        
        \multirow{2}{*}[-3pt]{Method} & 
        \multicolumn{2}{c|}{\textbf{Property}} &
        \multicolumn{6}{c|}{\textbf{F1-privacy (Prot / Percept.)}} &
        \multicolumn{3}{c|}{\textbf{F1-privacy (Prot / Det)}} & \multirow{2}{*}[-3pt]{Avg.}\\
        
         \cmidrule(lr){2-3} \cmidrule(lr){4-12} 
         & Train-free & General &VGG/SSIM & FN/SSIM & AF/SSIM & VGG/PSNR & FN/PSNR & AF/PSNR & VGG/Det & FN/Det & AF/Det  &\\
        %  &\multicolumn{3}{c|}{{$\text{VGGFace}\uparrow$}} & \multicolumn{3}{c|}{$\text{FaceNet}\uparrow$} & \multicolumn{3}{c}{{$\text{ArcFace}\uparrow$}} \\
        % \midrule
        % &&&&SSIM $\uparrow$ & $\text{PSNR}_{\text{N}} \uparrow$ & $\text{LPIPS}_{\text{N}} \uparrow$ & SSIM $\uparrow$ & $\text{PSNR}_{\text{N}} \uparrow$ & $\text{LPIPS}_{\text{N}} \uparrow$ & SSIM $\uparrow$ & $\text{PSNR}_{\text{N}} \uparrow$ & $\text{LPIPS}_{\text{N}} \uparrow$ \\
        \midrule

        % Method & 
        RIDDLE~\cite{riddle}     & \ding{55}& \ding{55}& 0.642 & 0.655 & 0.653 & 0.400 & 0.406 & 0.405 & \textbf{0.936} & \textbf{0.965} & \textbf{0.961} & 0.669 \\
        FALCO~\cite{falco}       & \ding{55} & \ding{55} &  0.696 & 0.690 & 0.712 & 0.543 & 0.539 & 0.552 & \underline{0.886} & 0.876 & 0.911 & 0.712 \\
        Simple~\cite{face_simple} & \ding{55} & \ding{55} &0.833 & 0.822 & 0.842 & 0.788 & \underline{0.779} & 0.796 & 0.862 & 0.851 & 0.871 & 0.827\\
        G2Face~\cite{g2face}     & \ding{55}& \ding{55}  &  \underline{0.853} & \underline{0.823} & \textbf{0.899} & \underline{0.797} & 0.771 & \textbf{0.838} & 0.878 & 0.846 & \underline{0.927} & \underline{0.848}\\
        \midrule
        Nullface~\cite{nullface} & \textcolor{myDeepG}{\ding {51}} & \ding{55}  & 0.698 & 0.602 & 0.633 & 0.664 & 0.576 & 0.605 & 0.713 & 0.613 & 0.646 & 0.639 \\
        \rowcolor{blue!5}
        APD & \textcolor{myDeepG}{\ding {51}}  & \textcolor{myDeepG}{\ding{51}} & \textbf{0.861} & \textbf{0.865} & \underline{0.881} & \textbf{0.808} & \textbf{0.811} & \underline{0.825} & 0.879 & \underline{0.883} & 0.900 & \textbf{0.857} \\
        \bottomrule
    \end{tabular}}
\vspace{-10pt}
\end{table*}

\noindent \textbf{MLLM Evaluation.} As traditional evaluation methods struggle with the complexities of composite privacy scenes, we adopt an MLLM-assisted assessment. MLLM evaluation covers the composite categories in AdaptShield and several relatively challenging textual categories, such as Receipt, which often involve complex content structures that pose difficulties for conventional OCR-based textual detection. As illustrated in Fig.~\ref{fig:mllm-example}, high fidelity methods like FluxEdit~\cite{fluxedit} and Instructp2p~\cite{icedit} still leak sensitive details (\textit{e.g.,} the name “Terry K Blodgett”), whereas strongly protected baselines produce heavily distorted results with low fidelity (\textit{e.g.,} Silhouette: 1.0). In contrast, our method achieves minimal fidelity loss (3.67, matching the original) while generating non-identifiable responses, ensuring effective privacy protection without compromising visual or semantic quality. Quantitatively, as shown in Tab.~\ref{tab:privacy_results_main}, our method achieves the highest $\text{F1-privacy}$ score on average across all categories and four MLLM evaluators, demonstrating that it preserves high fidelity ($F$) while effectively confusing the MLLMs with strong protection ($P$). Across all four evaluated MLLM series, our method achieves an average improvement of 8.5\% over the best baseline methods.
Notably, the largest gains are observed on Qwen3-VL-8B-Instruct and InternVL3.5-14B, with improvements of 10.8\% and 11.9\%, respectively.

\begin{wraptable}[17]{r}{0.54\linewidth}
% \vspace{-12pt}
\centering
\small
\setlength{\tabcolsep}{4pt}
\caption{
Ablation results under different parameters ($\eta$ and $\tau$), showing face similarity and image quality metrics. SSIM and PSNR evaluate preservation of sensitive regions; higher values indicate stronger structural consistency but greater identity leakage risk.
}
\vspace{-6pt}
\resizebox{\linewidth}{!}{
\begin{tabular}{c|c|ccc|cc}
\toprule
\multirow{2}{*}{Para.} & \multirow{2}{*}{Value}& \multicolumn{3}{c|}{\textbf{Concealment$\uparrow$}} & \multicolumn{2}{c}{\textbf{Consistency$\uparrow$}} \\
\cmidrule(lr){3-5} \cmidrule(lr){6-7} 
& & VGGFace & Facenet & ArcFace & PSNR & SSIM \\
\midrule
\multirow{4}{*}{\textbf{$\eta$}} 
& 0.1 & 0.8443 & 0.8894 & 0.8772 & 19.97 & 0.7550 \\
& 0.3 & 0.8110 & 0.8355 & 0.8395 & 21.58 & 0.7781 \\
& 0.5 & 0.7598 & 0.7584 & 0.7821 & 23.54 & 0.8070 \\
& 0.7 & 0.5957 & 0.5367 & 0.6081 & 26.21 & 0.8464 \\
\midrule
\multirow{4}{*}{\textbf{$\tau$}} 
& 5  & 0.7955 & 0.8275 & 0.8164 & 21.71 & 0.7820 \\
& 10 & 0.7398 & 0.7303 & 0.7421 & 23.46 & 0.8065 \\
& 15 & 0.6826 & 0.6328 & 0.6852 & 24.79 & 0.8254 \\
& 20 & 0.6058 & 0.5605 & 0.6257 & 26.56 & 0.8508 \\
\bottomrule
\end{tabular}
}
\vspace{-36pt}
\label{tab:ablation-eta-gamma-step}
\end{wraptable}
\noindent\textbf{Face Anonymization.} The efficacy of face anonymization acts as a crucial indicator for validating the capability of any technique designed for visual privacy protection.
As shown in Tab.~\ref{tab:face_tradeoff_scores}, we evaluate the F1-Privacy score across three recognition backbones (VGGFace, FaceNet, and ArcFace) for concealment, and perceptual metrics (SSIM, PSNR) together with face detection rate for fidelity.
Despite being entirely training-free and designed for generalization across diverse privacy categories, our method still achieves the best average score under all evaluation settings. This superiority is most evident in the $34.1\%$ improvement over another training-free method NullFace ($0.857$ vs. $0.639$). Although RIDDLE and Falco achieve competitive results in facial privacy protection, they substantially degrade background consistency, as shown in Tab.~\ref{tab:face_tradeoff_scores} and Fig.~\ref{fig:compares}. In contrast, our method better balances privacy protection and visual fidelity. Furthermore, unlike previous approaches that are limited to fixed identity replacement, our framework supports controllable facial attribute edits (\textit{e.g.,} age, gender, ethnicity), enabling user-tailored protection and greater adaptability for MLLM applications.

\subsection{Ablation Study}
% We conducted ablation studies to analyze the impact of the key control parameters $\eta$ and $\tau$, which influence the effect of Source Anchoring. 
% These essential parameters respectively control the duration (${\tau}$) and the magnitude (${\eta}$) of the source influence, making their combined control vital for achieving the optimal Protection-Fidelity trade-off during the editing process.

\begin{wrapfigure}[11]{r}{0.55\linewidth}
% \vspace{15pt}
\centering
\includegraphics[width=\linewidth]{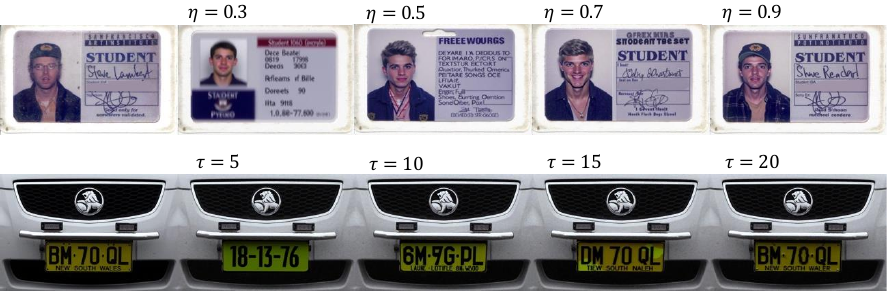}
% \vspace{-12pt}
\caption{
Visualization of results under different $\eta$ (upper row) and $\tau$ (bottom row).
}
% \vspace{-30pt}
\label{fig:ablation}
\end{wrapfigure}

\noindent \textbf{Impact of Edit Magnitude $\boldsymbol{\eta}$.}
The parameter $\eta$ defines the mixing ratio of influence between the source constraint and the privacy-preserving guidance during the editing process.
When $\eta$ is too small, structural details are weakened, occasionally resulting in distorted or implausible faces. Conversely, excessively large $\eta$ values make the generated face overly similar to the original, reducing anonymization effectiveness.
As shown in Fig.~\ref{fig:ablation} and Tab.~\ref{tab:ablation-eta-gamma-step}, 
larger $\eta$ improves structural preservation while retaining privacy protection. 
% we use $\eta=0.5$ for faces and $\eta=0.7$ otherwise.\\
% increasing $\eta$ gradually improves structural preservation while maintaining privacy protection. 
% We set $\eta=0.5$ for face images and $\eta=0.7$ for others. \\
% We set $\eta$ to 0.5 for faces and 0.7 for other images. \\

\textbf{Impact of Anchoring Timestep $\boldsymbol{\tau}$.}
The parameter $\tau$ is the anchoring timestep that defines the duration for which the source image guides the reverse diffusion process. Critically, $\tau$ controls the extent of the source's influence over the editing trajectory by defining the length of the guidance period.
If $\tau$ is too small, the guidance is removed too early, causing the sampling trajectory to drift from the source and degrading content fidelity.
Conversely, an excessively large $\tau$ keeps the guidance active for too long, making the result overly similar to the original and weakening anonymization.
% We set $\tau=10$ for face images and $\tau=15$ for others.

The choice of $\eta$ and $\tau$ depends on whether the privacy type involves explicit attribute substitution or relies on implicit stochastic divergence. Explicit substitutions (e.g., redirecting age or gender of faces) require more drift freedom for transformation, so we relax the anchoring with $\eta=0.5 $ and $\tau=10$. For implicit cases, where stochastic divergence already suffices,  we apply $\eta=0.7 $ and $\tau=15$ .

% \paragraph{Impact of $\gamma$}
% The parameter $\gamma$ controls the noise magnitude during inversion (higher $\gamma$ means stronger perturbation).
% A too-small $\gamma$ injects insufficient noise, keeping the inverted latent overly close to the source and resulting in poor anonymization.
% In contrast, larger $\gamma$ introduces excessive randomness, leaving limited room for controlled reconstruction. As shown in Tab.~\ref{tab:ablation-eta-gamma-step}, $\gamma$ exhibits low sensitivity within a moderate range, and we adopt $\gamma=0.9$ for our experiment.

\section{Related Work}
\textbf{Multimodal Large Language Model (MLLM).} Multimodal large language models (MLLMs) extend large language model (LLM) architectures to process and reason over vision (and sometimes other) modalities, enabling unified understanding, generation, and editing across image-text inputs. Early efforts such as MiniGPT‑4~\cite{minigpt4} align a frozen vision encoder with a strong LLM for rich image-driven text outputs and emergent image understanding. LLaVA~\cite{llava} and its successors use instruction-tuned vision-language adapters on top of base LLMs (\textit{e.g.,} LLaMA~\cite{llama}) to support tasks like visual question answering, image captioning, and interactive vision-language dialogues. More recent methods, such as SmartEdit~\cite{smartedit} and Step1X~\cite{step1x}, advance fine-grained instruction-based image editing, moving beyond pure understanding to generation and manipulation. Together, these works point toward the emergence of MLLMs capable of understanding, interpreting, and processing increasingly complex multimodal content in response to diverse and rich instructions.

\textbf{Privacy Protection for MLLMs.}
Privacy in MLLMs has attracted increasing attention, with recent work ranging from risk characterization to mitigation. Early approaches, such as Differential Privacy (DP) \cite{dp1,dualpriv}, limited what can be inferred from model outputs, while query-level methods like ReVision \cite{revision} and MARRS \cite{marrs} rewrite user queries or restructure inference to prevent exposure of raw visual data. Benchmarks such as MLLMU-Bench \cite{liu2025mllmu} evaluate MLLMs’ ability to unlearn sensitive knowledge. Studies by Chen \textit{et al.} \cite{chen2025mmprivacy} categorize privacy leakage into task-specific vulnerabilities, Zhang and Cheng \cite{zhang2025geo} highlight geolocation and profiling risks from visual cues, and Lin \textit{et al.} \cite{lin2025third} show that current systems often fail to identify sensitive contexts in smartphone agents. Collectively, these findings underscore the urgent need to balance privacy protection with visual fidelity.

\section{Conclusion}
% The widespread deployment of MLLMs and the rapid growth of user participation introduced new risks from diverse visual input carrying heterogeneous privacy leakage. 
The widespread deployment of MLLMs and surging user engagement have introduced novel risks, as diverse visual inputs often harbor heterogeneous privacy threats.
In this context, we presented a new paradigm for generalized visual privacy protection. Our proposed APD casts privacy protection as a controllable trajectory in latent space, drifting away from sensitive content while anchoring to the source for contextual fidelity. 
% making it well-suited to address the complexities of privacy protection in the MLLM era. 
Moreover, we developed AdaptShield to address the lack of comprehensive evaluation protocols for jointly evaluating privacy concealment and content fidelity. 
 Covering 22 heterogeneous privacy categories and 
employing four representative MLLM evaluators, AdaptShield establishes a systematic evaluation protocol and offers a standardized foundation for future research on privacy-aware multimodal intelligence. Experimental results demonstrate that this framework achieves superior efficacy and high adaptability across heterogeneous categories, highlighting its capacity to handle the complexities of privacy protection in the MLLM era.

\clearpage
\newpage
\bibliographystyle{plainnat}
\bibliography{main}

%%%%%%%%%%%%%%%%%%%%%%%%%%%%%%%%%%%%%%%%%%%%%%%%%%%%%%%%%%%%
\clearpage

\end{document}